\pdfoutput=1
\documentclass[11pt]{article}

\usepackage[preprint]{acl}

\usepackage{times}
\usepackage{latexsym}

\usepackage[T1]{fontenc}
\usepackage[utf8]{inputenc}

\usepackage{microtype}

\usepackage{inconsolata}

\usepackage{graphicx}

\usepackage{amsmath,amssymb}
\usepackage{hhline}
\usepackage{enumitem}
\usepackage{ltablex}
\usepackage{hyperref}
\usepackage{xcolor}
\usepackage{times}
\usepackage{latexsym}
\usepackage{xcolor,colortbl}
\usepackage{hyperref}
\usepackage{multirow}
\usepackage{tablefootnote}
\usepackage{mathtools}
\usepackage{amsfonts}
\usepackage{amsmath,amssymb}
\usepackage{hhline}
\usepackage{enumitem}
\usepackage{ltablex}
\usepackage{arydshln}
\usepackage{url}
\usepackage{array}
\usepackage{color}
\usepackage{tcolorbox}
\usepackage{tikz}
\usepackage{amssymb}

\newcolumntype{P}[1]{>{\raggedright\arraybackslash}p{#1}}

\usepackage[normalem]{ulem}
\usepackage{soul}
\usepackage{tabularx}

\definecolor{azure}{rgb}{0.0, 0.5, 1.0}
\tcbuselibrary{skins}
\tcbset{enhanced}

\usepackage{color}
\usepackage{tcolorbox}
\usepackage{tikz}
\usepackage{amssymb}

\definecolor{azure}{rgb}{0.0, 0.5, 1.0}
\definecolor{question-colour}{HTML}{e7f1ff}
\definecolor{section-colour}{HTML}{9ec6ff}
\definecolor{border-colour}{HTML}{0065f3}

\tcbuselibrary{skins}
\tcbset{enhanced}

\newcommand{\qsecboxqtn}[1]{\begin{tcolorbox}[colback=question-colour,colframe=border-colour]\raggedright{#1}\end{tcolorbox}}

\usepackage{wasysym}
\newcommand{\radiobutton}[0]{\ooalign{\hidewidth\cr$\ocircle$}}
\newcommand{\checkbox}[0]{\ooalign{\hidewidth\cr$\square$}}

\title{QRA++: Quantified Reproducibility Assessment for\\Four Common Types of Evaluation Results in NLP}
\title{QRA++: Quantified Reproducibility Assessment for\\Four Common Types of  Results in Natural Language Processing}
\title{QRA++: Quantified Reproducibility Assessment for\\Common Results Types in Natural Language Processing}
\title{QRA++: Quantified Reproducibility Assessment for\\Common Types of Results in Natural Language Processing}

\author{Anya Belz \\
  ADAPT Research Centre \\
  Dublin City University \\
  Dublin, Ireland \\
  \texttt{anya.belz@dcu.ie} \\}

\begin{document}
\maketitle
\begin{abstract}
Reproduction studies reported in NLP 
provide individual data points which in combination indicate worryingly low levels of reproducibility in the field. Because each reproduction study reports quantitative conclusions based on its own, often not explicitly stated, criteria for reproduction success/failure, the conclusions drawn are hard to interpret, compare, and learn from. In this paper, we present QRA++, a quantitative approach to reproducibility assessment that 
(i)~produces continuous-valued \textit{degree of reproducibility} assessments at three levels of granularity; 
(ii)~utilises reproducibility measures that are directly \textit{comparable} across different studies; and (iii) grounds  expectations about degree of reproducibility in degree of similarity between experiments. 
QRA++ enables more informative reproducibility assessments to be conducted,
and conclusions to be drawn about what causes reproducibility to be better/poorer. 
We illustrate this by applying QRA++ to three example sets of comparable experiments, revealing clear evidence that degree of reproducibility depends on similarity of experiment properties, but also system type and evaluation method.
\end{abstract}

\section{Introduction and Background}\label{sec:intro}

A number of reproduction results have accumulated in Natural Language Processing where one set of researchers repeated (at least the evaluation of) research conducted by another set of researchers, either (i) as part of a reproduction study, or (ii) in the course of other research, typically to enable fairer\footnote{I.e.\ fairer than comparing against results reported in papers which are mostly not directly comparable.} comparison of a new system to an existing one. For example, \citet{belz-etal-2021-systematic} collected 513 pairs of scores produced in the latter context, nearly all computed with metrics. Five years of shared tasks on reproduction studies have between them produced several hundred more score pairs \cite{branco-etal-2020-shared,
%belz2021reprogen,belz-etal-2022,belz-thomson-2023-2023,
belz20242024}, nearly all for human evaluations. From such paired results, we know that repeated evaluations rarely produce the same scores even for automatically computed metrics. However, that is almost all we know. Conclusions and findings in such comparisons are usually qualitative and narrative in nature, involving binary statements of whether results have been reproduced (or not), or whether reproduction has been successful (or not). 

What we lack is a solid \textit{quantitative} basis on which repeat experiments and evaluations can be assessed for their \textit{degree} of reproducibility (instead of binary qualitative conclusions which are hard to generalise and learn from), moreover in a way that is \textit{comparable} across assessments. 
In order to gain a complete picture of reproducibility, we need to be able to do this at different levels of granularity, not just for individual system score pairs (e.g.\ small-sample coefficient of variation; \citeauthor{belz2022metrological}, \citeyear{belz2022metrological}), but also for system rankings and conclusions from experiments. To provide these abilities is the aim of the Extended Quantified Reproducibility Assessment (QRA++) framework presented in this paper.

\section{Basics from Metrology}\label{sec:term-conc}

Following \citet{belz2022metrological}, we base our work on the general meta-scientific principles of metrology. The International Vocabulary of Metrology (VIM) \cite{jcgm2012international} defines repeatability and reproducibility as follows (defined terms in bold, see VIM for subsidiary defined terms):

\begin{enumerate}[topsep=4pt,itemsep=0pt,parsep=3pt,partopsep=0pt]
 \item[{2.21}] \textbf{measurement repeatability} (or repeatability, for short) is  \textbf{measurement  precision}  under  a  set  of  \textbf{repeatability conditions of measurement}.   

\item[{2.20}] a \textbf{repeatability condition of measurement} (repeatability condition) is a condition  of  \textbf{measurement},  out  of  a  set  of  conditions   that   includes   the   same   \textbf{measurement procedure},   same   operators,   same   \textbf{measuring system},   same   operating   conditions   and   same   location,  and  replicate  measurements  on  the  same  or similar objects over a short period of time. 

\item[{2.25}] \textbf{measurement reproducibility} (reproducibility) is \textbf{measurement   precision}   under   \textbf{reproducibility conditions of measurement}. 

\item[{2.24}] a \textbf{reproducibility condition of measurement} (reproducibility condition) is a condition  of  \textbf{measurement},  out  of  a  set  of  conditions  that  includes  different  locations,  operators,  \textbf{measuring  systems}, etc.  
A   specification   should   give   the   conditions   changed and unchanged, to the extent practical.
\end{enumerate} 

\noindent The above definitions mean that repeatability and reproducibility are properties \textit{not} of objects, scores, results or conclusions (as commonly assumed in NLP/ML), but of measurements $M_i$ which are parameterised by measurand $m$, object $O$, time $t_i$, and conditions of measurement $C_i$ (just $C$, i.e.\ the same for all $M_i$ in the case of repeatability), and return a measured quantity value $v_i$:

\vspace{-.4cm}
%\begin{small}
\begin{equation}
%\begin{aligned}
\begin{split}
R(M_1,\! M_2,\! ... M_n) := \text{Precision}(v_1, \!v_2, \!... v_n),\\ \text{where }
M_i: (m, O, t_i, C_i) \mapsto v_i\\
\end{split}
%\end{aligned}
\label{def:repro}
\end{equation}
%\end{small}
\vspace{-.3cm}

\noindent Repeatability and reproducibility, jointly denoted $R$ above, are defined as measurement precision, and are quantified by calculating the precision of a set of measured quantity values relative to a set of conditions of measurement $C_i$
which have to be known and specified for assessment of 
repeatability and reproducibility to be meaningful. Repeatability is a special case of reproducibility where $C_i$ is the same for all $M_i$.

\section{Terminology, Desiderata, Components}\label{sec:basic-frame}

\subsection{Terminological conventions}\label{ssec:terminology}

Mapping the terms from metrology in the previous section to NLP/ML, 
an \textit{object} is an \textbf{system}, a \textit{measurand} is an \textbf{evaluation measure}, and the \textit{measurement conditions} are \textbf{experiment properties}. Under repeatability conditions of measurement, all experiment properties are the same in all runs of the experiment, while under reproducibility conditions, some may differ. 
We use the term \textbf{experiment} to refer to an assessment of a single set of systems on a single \textit{evaluation measure}. The term \textbf{study} refers to a set of experiments conducted jointly.

In many cases, we can distinguish an \textbf{original experiment} and a later \textbf{repetition} (\textbf{re-run}, \textbf{reproduction study}) of it. This is not always the case and we use the term \textbf{comparable experiments} where the distinction is unimportant. 

In a typical NLP scenario, we build some new systems and take some existing systems for comparison, plus baselines and possibly toplines, then evaluate them using some evaluation method, producing a set of quantity values that represent the outcome of evaluation. In metrological terms, all of the above, apart from the systems (objects) themselves, constitutes a \textit{measurement} that produces a \textit{measured value} for each object. We use the terms \textbf{evaluation} and \textbf{evaluation result} to refer to these two items in an NLP/ML recognisable way. 

Recreation of systems is also sometimes addressed in NLP under the heading of reproducibility, but while the framework presented here can straightforwardly be extended to encompass this (e.g.\ as in \citeauthor{belz2022metrological}, \citeyear{belz2022metrological}, for metrics) we do not directly discuss this aspect in the present paper.

\begin{table*}[h!t]
    \centering\small
    \begin{tabular}{ll}

\begin{tabular}{|p{7cm}|}
\hline
\multicolumn{1}{|c|}{General evaluation experiment properties} \\
\hline     
{Test dataset.} \\ 
Metric.\\
Metric implementation.\\
Run-time environment.\\
{H2.1 Input type}.\\
{H2.2 Output type}.\\
{H2.3 Task}.\\
{H3.1.1 Total evaluated items}.\\
{H4.2.1 Objective vs.\ subjective evaluation mode}.\\
{H4.2.2 Absolute vs.\ relative evaluation mode}.\\
{H4.2.3 Intrinsic vs.\ extrinsic evaluation mode}.\\
\hline
\end{tabular}

&

\begin{tabular}{|p{7cm}|}
\hline
\multicolumn{1}{|c|}{Human evaluation experiment properties} \\
\hline 
{H3.2.1 Number of evaluators}.\\
{H3.2.2.1 Evaluator domain expertise}.  \\
{H3.2.2.4 Authors among evaluators}.\\
{H3.2.4 Evaluator training/practice}.\\
{H3.2.5 Evaluator type}.\\
{H3.3.2 Response collection tool/platform}.\\
{H3.3.3.1 Quality assurance methods}.\\
{H4.3.1.2 Standardised quality criterion}.\\
{H4.3.5 Rating instrument type}.\\
{H4.3.7 Verbatim instrument question or prompt}.\\
{H4.3.8 Form of response elicitation}.\\
\hline
\end{tabular}

\\

\end{tabular}
\caption{Experiment properties (measurement conditions) based on the HEDS 3.0 \cite{belz2024heds} questions and answers given in full in  Appendix~\ref{sec:appendix-exp-prop}.}\label{tab:exp-prop}
\end{table*}

\subsection{Desiderata}\label{ssec:desiderata}

In NLP, reproducibility is often treated as a relationship between \textit{two} experiments, an original one and a re-run of it. However, we know from existing research \cite{belz20242024} that two re-runs of the same original experiment can produce very different reproducibility results. When that is the case, which set of results provides an indication of the true reproducibility of the experiment? The answer from metrology, adopted in QRA++, is that we need to produce reproducibility assessments on the basis of all three sets of results (original, re-run~1, re-run~2), or all \textit{n} in the general case.

This, in combination with the aspects mentioned in the preceding two sections, gives us the following desirable high-level properties for a quantitative reproducibility assessment framework. It should:

\vspace{-0.1cm}
\begin{itemize}\itemsep=-0.1pt
    \item Be designed to assess reproducibility on the basis of two or more comparable studies;
    \item Be grounded in a property-based definition of similarity between experiments to {(a)} guide expectations of reproducibility, and {(b)} support conclusions about which properties of experiments are associated with better, and which with worse, reproducibility;
    \item Produce continuous-valued assessments to capture degree of reproducibility (not binary outcomes);
    \item Support assessment at multiple levels of granularity (even if scores  differ, overall conclusions can still be the same); and
    \item Provide automatically computable degree of reproducibility measures for different types of results (e.g.\ labels need to be handled differently than metric scores). 
\end{itemize}

\subsection{Components}\label{ssec:components}

To address the above desiderata, QRA++ has the following main components: (i) a standard \textbf{\textit{set of experiment properties}} (measurement conditions) for capturing the degree of similarity between experiments: important because we would expect more similar experiments to have better reproducibility; (ii) four \textbf{\textit{types of experiment results}}: depending on result type, different measures of reproducibility are appropriate; (iii) four corresponding \textbf{\textit{sets of reproducibility measures}} for computing the degree of reproducibility of \textit{two or more} aligned sets of results obtained from comparable experiments (repeat measurements), at the \textit{system, quality criterion, and study levels}; and (iv) a \textbf{\textit{reporting template}} for ease of comparison between reproducibility assessments. We describe each of these components in turn in Sections~\ref{sec:exp-prop}, \ref{sec:exp-results}, \ref{sec:repro-measures}, and \ref{sec:recording-template}.

\section{Experiment Properties}\label{sec:exp-prop}

Experiment properties have the role of measurement conditions (Section~\ref{sec:term-conc}). In strict reproducibility assessment (repeatability in Section~\ref{sec:term-conc}), all property values need to be the same for experiments to be comparable. 

The experiment properties listed in Table~\ref{tab:exp-prop} are a subset of 18 HEDS~3.0 questions \cite{belz2024heds}, here shortened and simplified, augmented by four properties not in HEDS~3.0 (top left). The list on the left contains properties agnostic about metric-based vs.\ human evaluation, the list on the right contains properties specific to human evaluation (see Appendix~\ref{sec:appendix-exp-prop} for full details of the 18 HEDS questions). Note that one of the agnostic properties, run-time environment, does not always apply in the case of human evaluations where there may be no code to execute at all.

We are using a subset of possible experiment properties here, because what we are after is the minimal set of properties for which, if they are the same, we would expect the outcome of evaluations to also be the same. In other words, if these properties are the same and outcomes differ then we may have a reproducibility problem, whereas if these properties differ, differences in outcomes are in fact expected (although not always seen).
This is not a distinction for which there is an objective basis (at least not without a substantially larger body of reproduction studies grounded in experiment properties than we currently have available to us), and it is more accurate to view the selected set of experiment properties as a way of controlling the strictness of the reproducibility assessment.

The properties also allow systematically varying certain aspects of experiments for their effect on reproducibility, to study e.g.\ what happens when we swap out the test dataset or participant type.

\section{Types of Experiment Results}\label{sec:exp-results}

QRA++ distinguishes four main types of results commonly reported in NLP and ML papers: 

\vspace{-.2cm}\begin{enumerate}\itemsep=-.1cm
    \item Type I results: single numerical scores, e.g.\ BLEU, mean quality rating, error count, etc.
\item Type II results: sets of related numerical scores, e.g.\ a set of Type I results. 
\item Type III results: categorical labels attached to text spans of any length, e.g.\ error annotations, outputs classified in some way.
\item Type IV results: Findings relating to differences between systems, 
e.g.\ System~A is significantly better than System~B in terms of a given evaluation method and test metric.
\end{enumerate}
\vspace{-.2cm}

\begin{table*}[t]
\setlength\tabcolsep{4pt} % default value: 6pt
\renewcommand{\arraystretch}{1.25} % Default value: 1
    \centering\small
    \begin{tabular}{|p{1cm}|p{3.7cm}||p{0.8cm}|P{1.3cm}||p{2cm}|P{2.6cm}|P{2.35cm}|}
\hline
        &  & \multicolumn{5}{c|}{\textbf{Degree of reproducibility measure(s)}}\\
\cline{3-7}
       Type of Result & Description & \multicolumn{2}{p{2.1cm}||}{$n$ experiments being compared} & \multicolumn{3}{c|}{Level at which reproducibility assessed}\\
       \cline{3-7}
     &  & \centering$n=2$ & \centering $n>2$ & \centering System & \centering Quality Criterion & \hspace{0.5cm} Study \\
\hline
\hline
       \multirow{3}{*}{Type I}  & Single numerical scores, e.g.\ BLEU score, mean quality rating, error count, etc. & \multirow{3}{*}{CV$^*$} & \multirow{3}{*}{CV$^*$} & \cellcolor{blue!20}\multirow{3}{*}{CV$^*$} & \multirow{2}{2.6cm}{mean of all system- level CV$^*$ scores for QC} & \multirow{2}{2.35cm}{mean of all system-level CV$^*$ scores for study}\\
\hline
       \multirow{2}{*}{Type II}  & Sets of related numerical scores, e.g.\ a set of Type I results. & \multirow{3}{*}{$r, \rho, \tau$} & mean pair- wise $r$, $\rho$; $W$ & \multirow{3}{*}{n/a} & \cellcolor{blue!20} (mean) $r, \rho$; $\tau$ or $W$ computed over all scores for QC  & \multirow{3}{*}{n/a}\\
\hline
       \multirow{4}{*}{Type III}  & Categorical labels attached to text spans of any length, e.g.\ error annotations, outputs classified in some way. & \multirow{3}{*}{$\kappa$} & \multirow{3}{*}{$\alpha$} & $\kappa$, $\alpha$ computed over all labels for system & \cellcolor{blue!20}$\kappa$, $\alpha$ computed over all labels for QC & mean of QC-level $\kappa$, $\alpha$ scores\\ 
\hline
       \multirow{4}{*}{Type IV}  & Findings re differences between systems stated explicitly or implied by quantitative results in the original paper. 
       & \multirow{3}{*}{$P$} & \multirow{3}{*}{$P$} & \multirow{3}{*}{n/a} & $P$ computed over all differences for same QC & \cellcolor{blue!20}$P$ computed over all differences in study\\
\hline
       
    \end{tabular}\caption{QRA++ results types alongside corresponding degree of reproducibility measures and level at which they operate; $r$ = Pearson's $r$, $\rho$ = Spearman's $\rho$, $\tau$ = Kendall's $\tau$, $W$ = Kendall's $W$, $\kappa$ = Fleiss’s $\kappa$; $\alpha$ = Krippendorff’s $\alpha$. 
    n/a = measure(s) cannot be applied at this level; $P$ = proportion of differences between pairs of systems that have the same sign (see Section~\ref{fig:P-definition}).}
    \label{tab:res-types}
\end{table*}

\noindent Table~\ref{tab:res-types} provides an overview of the four types of results, each alongside corresponding measures for assessing degree of reproducibility, the latter as introduced in the next section.

\section{Reproducibility Measures}\label{sec:repro-measures}

QRA++ reproducibility measures come in two flavours (columns~3 and~4 in Table~\ref{tab:res-types}): (i) the case where we have \textbf{\textit{two comparable experiments}} (typically an original experiment and a reproduction of it); and (ii) the case where we have \textbf{\textit{more than two comparable experiments}} which ideally we should have for more reliable assessments of reproducibility but which is extremely hard to achieve in NLP.

Moreover, 
degree-of-reproducibility measures can usefully be computed at three different levels (columns~5--7 in Table~\ref{tab:res-types}): (i) the \textbf{system level}, where scores for the same individual system are compared (across two or more studies); (ii) the \textbf{quality criterion (QC) level}, where scores for all systems assessed in terms of a given single QC are compared; and (iii) the \textbf{study level}, where scores for all systems and all QCs in a study are compared. Some of the degree-of-reproducibility measures extend natively (without adjustment to the measure, only to the set of scores compared) from one level to the next; if that is not the case, we use a generalisation of the measure where one exists, or compute the mean of measures at the applicable level. 

We assess the four results types quantitatively by the following degree of reproducibility measures, each described in more detail over the following four subsections):

\vspace{-.25cm}    
\begin{enumerate}\itemsep=-.2pt
    \item Type I results: Small-sample coefficient of variation CV$^*$ \cite{belz2022metrological}.
    \item Type II results: Pearson’s $r$, Spearman’s $\rho$, Kendall's $\tau$, Kendall's $W$.
    \item Type III results: Multi-rater: Fleiss’s $\kappa$; Multi-rater, multi-label: Krippendorff’s $\alpha$.
    \item Type IV results: Proportion of all pairwise system ranks that are the same in two experiments. 
    \end{enumerate}
\vspace{-.25cm}    

\noindent Note that each degree of reproducibility measure has a level at which it most meaningfully applies which we mark in Table~\ref{tab:res-types} with blue shading. For example, CV$^*$ and mean $\kappa$, $\alpha$ over QCs can be computed at the study level but will completely obscure meaningful, and potentially big, differences between quality criteria. This matters: in the case of CV$^*$, we know score-level similarity can be very different for different QCs in the same study, while in the case of $\kappa$ and $\alpha$, labelling similarity can differ widely depending on label sets.

\subsection{Type I results}\label{ssec:type-I}

Following \citet{belz2022metrological}, we use CV$^*$ for our standard precision measure. In other fields, measurement precision is typically reported in terms of the  coefficient of variation (CV), i.e.\ standard deviation $s$ over mean $m$. 
CV serves well as the ‘headline’ result, because it is a general measure, not in the unit of the measurements (unlike mean and standard deviation), providing a quantification of degree of reproducibility that is comparable across studies \cite[p.\ 57]{ahmed1995pooling}. 

As reproduction studies in NLP/ML mostly have very small sample sizes $n$ of 2--4, standard CV can produce overly optimistic results. We address this by (i) applying a small sample correction $(1+\frac{1}{4n})$ \cite{sokal:rohlf:1969} to the overall CV; and (ii) using the {unbiased sample standard deviation}, denoted $s^*$. Combined, this gives the
\textbf{unbiased coefficient of variation CV$^*$} as:

\vspace{-.2cm}

%\begin{small}
\begin{equation}
\begin{aligned}
%\begin{split}
& \text{CV}^* =  \,\,\Big(1+\frac{1}{4n}\Big)\frac{\,\,s^*}{|m|}  \\
%& \text{where } s^* \, \text{ is the unbiased standard deviation, } \\
& \text{where } n \, \text{ is the sample size,}\\ 
\vspace{-.5cm}& \text{and } m \text{ the sample mean.}
%\end{split}
\end{aligned}
%\label{def:repro}
\end{equation}
%\end{small}
%\vspace{-.6cm}

\noindent We obtain $s^*$ by applying a standard bias correction for the case of normal distribution, where we divide the sample standard deviation $s$ by a correction factor $c_4(n)$ \cite{rao1973linear}:\footnote{For completeness, the sample standard deviation $s$ is given by 
$s = \sqrt{\frac{\sum_{i=1}^{n} (v_i-\Bar{v_i})^2}{n-1} }$, i.e.\ the square root of the variance computed with $n-1$ instead of $n$ (Bessel's correction). While the estimation of the population variance is then unbiased, the estimation of the population standard deviation still contains bias which is addressed with the $c_4(n)$ correction described.}

\vspace{-.2cm}
\begin{equation}
\begin{aligned}
& s^* = \frac{s}{c_4(n)} \\
& c_4(n) = \sqrt{\frac{2}{n-1}} \frac{\Gamma(\frac{n}{2})}{\Gamma(\frac{n-1}{2})} \\
\end{aligned}
\end{equation}

\noindent Confidence intervals for $s^*$ are calculated using a t-distribution, and the standard error of  $s^*$ is approximated on the basis of the standard error of the unbiased sample variance $\text{se}(s^2)$ as $\text{se}_{s^2}(s^*) \approx \frac{1}{2\sigma}\text{se}(s^2)$ \cite{rao1973linear}. Assuming normal distribution, the standard error of the sample variance is calculated in the usual way as $\text{se}(s^2) = \sqrt{\frac{2\sigma^4}{n-1}}$. 

Before applying CV$^*$ to values on scales that do not start at 0 (in NLP this happens mostly in human evaluations) values need to be shifted to start at 0 to ensure comparability.\footnote{Otherwise CV$^*$ reflects differences solely due to different minimum values.}  Moreover, following convention in using CV for computing precision \cite{aahfws:nd}, we report CV$^*$ values scaled up by a factor of 100 (i.e.\ as percentages).

On the basis of reports for CV$^*$ values over the past few years (e.g.\ from the ReproNLP shared task series, \citeyear{belz-thomson-2023-2023,belz20242024}), values in the range below about 12 for human evaluations, and below 1 for metric-based evaluations can be considered to indicate good system-level reproducibility in contexts where an identical set of system outputs (not regenerated) is assessed, and experiment properties (Section~\ref{sec:exp-prop}) are all the same.\footnote{What counts as good reproducibility differs depending on what is measured. E.g.\ in bio-science, CV ranges from <10\% for enzyme assays, to 20--50\% for in vivo and cell-based assays, and >300\% for virus titer assays \cite{aahfws:nd}.}

CV$^*$ measures reproducibility in the metrology sense of the term (as precision) at the system level (Table~\ref{tab:res-types}). At the QC and study levels, it can still be used to produce points of reference and comparison, but with the proviso that it completely obscures any differences between QCs which can be large (as we shall see in the example applications in Section~\ref{sec:recording-template}).

\subsection{Type II results}\label{ssec:type-II}

For Type~II results (sets of scores typically from an evaluation method used to compare a set of systems), we use three standard measures, each giving a slightly different perspective on correlation: Pearson's $r$, Spearman's $\rho$, and Kendall's $\tau$. 

These all natively assess the case of $n=2$, and Kendall's $\tau$ has a generalisation for $n>2$, namely Kendall's $W$. For Pearson's and Spearman's, we compute the mean of the pairwise values when $n>2$. $W$ and $\tau$ can be considered the headline measures for Type II results, but $r$ and $\rho$ can provide additional, perhaps more intuitive and familiar, reference points.

As regards the three assessment levels, Type~II measures don't apply at the system level (single scores) or study level (multiple sets of scores potentially on very different scales), and natively apply only at the QC level.

\begin{figure*}[t!]
    \centering\begin{small}
\begin{equation}
\begin{aligned}
%\begin{split}
P \;=  \;\;& \frac{1}{|\mathbb{E}| \,|\mathbb{S}|}\,\sum_{(E_i,E_j) \in \mathbb{E}}\,\sum_{(S_m,S_n) \in \mathbb{S}} \bigg\{{{{1 \text{ , if sgn}\big(M_{E_i}(S_m)-M_{E_i}}(S_n)\big) = \text{sgn}(M_{E_j}(S_m) - M_{E_j}(S_n) \big)} \atop{\;\,0 \text{ , otherwise}\hspace{7.1cm}}}\\
\text{ where } & \; \mathbb{E} = \big\{ ( E_i, E_j )\; \big| \;E_i, E_j \in \mathbf{E} \land E_i \neq E_j \big\} \text{ , }\\
& \; \mathbb{S} = \big\{ ( S_m, S_n )\; \big| \;S_m, S_n \in \mathbf{S} \land S_m \neq S_n \big\} \text{ , }\\
& \; \textbf{E} \text{ is a set of comparable experiments, }\\
& \; \textbf{S} \text{ is the set of systems evaluated in } \textbf{E} \text{, and }\\
& \; M_E(S) \text{ is the evaluation method } M \text{ applied to system } S \text{ in experiment } E \text{ . }
%\end{split}
\end{aligned}
%\label{def:repro}
\end{equation}
\end{small}
\vspace{-.4cm}
\caption{$P$, the proportion of identical pairwise system ranks in a set of comparable experiments $\mathbf{E}$.}
\label{fig:P-definition}
\end{figure*}

\subsection{Type III results}\label{ssec:type-III}

Type~III results are categorical labels attached to text spans of any length, including class labels attached to entire dataset items. When such approaches are used for producing an evaluation score, they need to be aggregated in one way or another (how this is done can make a difference to findings, see e.g.\ \citeauthor{popovic2022reporting}, \citeyear{popovic2022reporting}). Once aggregated at the system level, such results become Type~I results and their reproducibility is assessed as per Section~\ref{ssec:type-I}.

However, there are cases where it is of interest to compare the sets of labels from two comparable experiments in their raw state. Here we propose to use standard inter-rater agreement measures, namely Cohen's $\kappa$ for the case of $n=2$, and Krippendorff's $\alpha$ for $n>2$. In other contexts these measures typically assess (literally) the agreement between two raters and more than two raters, respectively. When using them to assess reproducibility, 
we consider all labels from an experiment as produced by one rater. This works at the system and QC level, but not natively at the study level (beccause each quality criterion is likely to have different label sets). At the latter level, the means of the measures can be used to produce points of reference.

\subsection{Type IV results}\label{ssec:type-IV}

Type~IV results are arguably the results we care most about in system comparison in NLP/ML, providing information regarding which system performs better than another on a given task. Findings can be, and are, stated in many different ways in papers which makes them difficult to compare. In order to put the degree of reproducibility of findings on an objective footing, we define a measure, $P$, to be equal to the \textbf{proportion of identical pairwise system ranks} in the given set of comparable experiments, as shown in Figure~\ref{fig:P-definition}. In the case where we have two comparable experiments, this amounts to counting the number of system pairs that are ranked identically in the two experiments, and normalising the count by the total number of system pairs. In the case of $n$ comparable experiments, we do this over all pairs of experiments. 

$P$ is not applicable at the system level, but applies natively 
to any number of studies being compared, and at both QC level and study level.

\begin{table*}[t]
    \centering\small
    \begin{tabular}{|p{2cm}|p{2.2cm}|p{1.8cm}|p{2.2cm}|p{1.85cm}|}
\hline
        &  & \multicolumn{3}{c|}{\textbf{Degree of reproducibility ($n=3$)}}\\
\cline{3-5}
       \textbf{Type of Result} & Measure applied &  System level & QC level & Study level \\
\hline
\hline
       \multirow{1}{*}{\textbf{Type I}}   & \multirow{4}{*}{(mean) CV$^*$} & \cellcolor{blue!20} &  & \\
       $\;\;$ SVM     &  & \cellcolor{blue!20}19.96 & \multirow{3}{*}{26.87} & \multirow{3}{*}{26.87} \\
       $\;\;$ GeDi    &  & \cellcolor{blue!20}29.9 &  &  \\
       $\;\;$ DExpert &  & \cellcolor{blue!20}30.76 &  &  \\
\hline
       \multirow{3}{*}{\textbf{Type II}}  & mean $r$ & n/a & \cellcolor{blue!20}0.97 & n/a\\
                                 & mean $\rho$ & n/a & \cellcolor{blue!20}1 & n/a\\
                                 & $W$ & n/a & \cellcolor{blue!20}1 & n/a \\
\hline
       \multirow{1}{*}{\textbf{Type IV}}  & $P$ & n/a & 1 & \cellcolor{blue!20} 1 \\
\hline
       
    \end{tabular}\caption{Example 1: QRA++ reproducibility assessment for three comparable experiments ($n=3)$ from ReproNLP 2024: QC = \textbf{\textul{Fluency}}; \citet{august-etal-2022-generating,repronlp_submission_9_vanmiltenburg,repronlp_submission_27_li}. 
    Study level and QC level columns same for Type I and IV, because only one QC taken into account. For measures, see Table~\ref{tab:res-types} and Section~\ref{sec:repro-measures}; n/a $=$ measure does not apply at this level.}
    \label{tab:example-1}
\end{table*}

\begin{table*}[t]
    \centering\small
    \begin{tabular}{|p{2cm}|p{2.2cm}|p{1.8cm}|p{1.85cm}|p{1.85cm}|}
\hline
        \multirow{2}{*}{\textbf{Type of Result}} & \multirow{2}{*}{\textbf{Measure applied}} & \multicolumn{3}{c|}{\textbf{Degree of reproducibility ($n=2$)}}\\
\cline{3-5}
        &  &  System level& QC level & Study level \\
\hline
\hline
       \multirow{1}{*}{\textbf{Type I}}   &  &  &  & \multirow{10}{*}{27.65}\\
       $\;$ \textit{QC 1:} & \multirow{4}{*}{(mean) CV$^*$} & & & \multirow{8}{*}{}\\
       $\;\;$ T5-base     &  & \cellcolor{blue!20}47.25 & \multirow{3}{*}{44.83} &  \\
       $\;\;$ T5-large    &  & \cellcolor{blue!20}54.72 &  &  \\
       $\;\;$ GPT2-large  &  & \cellcolor{blue!20}32.53 &  &  \\
%       \cdashline{1-4}
       $\;$ \textit{QC 2:} & \multirow{4}{*}{(mean) CV$^*$} & & & \\
       $\;\;$ T5-base     &  & \cellcolor{blue!20}9.18 & \multirow{3}{*}{10.46} &  \\
       $\;\;$ T5-large    &  & \cellcolor{blue!20}8.86 &  &  \\
       $\;\;$ GPT2-large  &  & \cellcolor{blue!20}13.34 &  &  \\
\hline
       \multirow{1}{*}{\textbf{Type II}}  & & & & \\
%       \hdashline
       $\;$ \textit{QC 1:} &  & & & \multirow{8}{*}{}\\
        & $r$ & n/a & \cellcolor{blue!20}-0.968 & n/a\\
        & $\rho$ & n/a & \cellcolor{blue!20}-1 & n/a\\
        & $\tau$ & n/a & \cellcolor{blue!20}-1 & n/a  \\
%       \cdashline{1-4}
       $\;$ \textit{QC 2:} &  & & & \\
        & $r$ & n/a & \cellcolor{blue!20}0.582 & n/a \\
        & $\rho$ & n/a & \cellcolor{blue!20}0.866 & n/a\\
        & $\tau$ & n/a & \cellcolor{blue!20}0.816 & n/a \\
%\hline
%       \multirow{1}{*}{\textbf{Type III}}  & -- & -- & -- & -- \\
\hline
       \multirow{1}{*}{\textbf{Type IV}}  &  &  &  &  \cellcolor{blue!20}\\
%       \hdashline
       $\;$ \textit{QC 1:} & \multirow{1}{*}{$P$} & n/a & 0 & {0.333\cellcolor{blue!20}}\\
%       \cdashline{1-4}
       $\;$ \textit{QC 2:} & $P$ & n/a & 0.667 & \cellcolor{blue!20}\\
\hline
       
    \end{tabular}\caption{Example 2: Results from QRA++ reproducibility assessment for two comparable experiments ($n=2)$ from ReproNLP 2024: QC~1 = \textbf{\textul{Overall Quality}}; QC~2 = \textul{\textbf{Sociopolitical Acceptability}} \cite{gabriel-etal-2022-misinfo,repronlp_submission_34_mahlaza}. For measures, see Table~\ref{tab:res-types} and Section~\ref{sec:repro-measures}; n/a $=$ measure does not apply at this level. %Type~III row marked `--' because there were no such results.
    }
    \label{tab:example-2}
\end{table*}

\section{Reporting Template and Examples}\label{sec:recording-template}

Reporting templates can be based in different ways on Table~\ref{tab:res-types}, selecting applicable results types and then completing each applicable row and column for the systems and quality criteria in the study. 

Tables~\ref{tab:example-1}, \ref{tab:example-2}~and~\ref{tab:example-3} illustrate one way of doing this, where the single resulting table reports on the reproducibility of the entire \textit{study}, i.e.\ including all QCs (corresponding to individual experiments in our definition, see Section~\ref{sec:basic-frame}). Here, the first column identifies the result type (I--IV) and, for Type~I and~III results where system-level reproducibility can be assessed, the systems that were evaluated. The second column identifies the QRA++ reproducibility measure applied. The remaining three columns report the values obtained for the measures, at the level of individual systems, each quality criterion, and the study as a whole (all QCs). %, as applicable. 
Finally, we identify the sample size in the joint header for the last three columns (e.g.\ in Table~\ref{tab:example-1}, $n=3$).

\subsection{Example 1: Fluency of scientific definit- ions in 3 comparable experiments}

Table~\ref{tab:example-1} reports QRA++ results based on three comparable experiments for one QC, Fluency: the original work reported by \citet{august-etal-2022-generating}, and two repeat runs of the experiment, by \citet{repronlp_submission_9_vanmiltenburg} and \citet{repronlp_submission_27_li}. The experiment properties from Section~\ref{sec:exp-prop} were exactly the same in all three experiments. Fluency was assessed on a 5-point rating scale for three different systems, so we have Type I, II, and IV results only. %, but not Type~III. 
We only have one QC, so there is no distinct study level.

Looking at reproducibility at the system level, CV$^*$ values can be considered medium good (see end of Section~\ref{ssec:type-I}). Because we have three comparable experiments, we compute mean $r$, $\rho$ (both range $[-1,1]$), and $W$ (range $[0,1]$), for Type~II results, all of which are exceptionally high, indicating (near) perfect correlation.  

Finally, $P$ is $1$, indicating that both repeat experiments obtained exactly the same three pairwise differences as the original experiment. 

Overall, the QRA++ assessment for these three comparable experiments shows the experimental design to have medium reproducibility for Type~I results (individual, system-level scores), and perfect or near perfect reproducibility for Type~II (all system-level scores for the same quality criterion) and Type~IV (findings) results.

\begin{table*}[t]
    \centering\small
    \begin{tabular}{|p{2.2cm}|p{2.2cm}|p{1.8cm}|p{2.2cm}|p{1.85cm}|}
\hline
        &  & \multicolumn{3}{c|}{\textbf{Degree of reproducibility ($n=8$)}}\\
\cline{3-5}
       \textbf{Type of Result} & Measure applied &  System level & QC level & Study level \\
\hline
\hline
       \multirow{1}{*}{\textbf{Type I}}   & \multirow{11}{*}{(mean) CV$^*$} & \cellcolor{blue!20} & \multirow{11}{*}{10.95} & \\
       $\;\;$ mult-base     &     & 14.34  \cellcolor{blue!20}  & \multirow{3}{*}{ } & \multirow{9}{*}{10.95} \\
       $\;\;$ mult-word-L-    &     &  10.33 \cellcolor{blue!20}  &  &  \\
       $\;\;$ mult-word-L+    &     &  9.91 \cellcolor{blue!20}  &  &  \\
       $\;\;$ mult-pos-L-     &     &  3.88  \cellcolor{blue!20}  &  &  \\
       $\;\;$ mult-pos-L+     &     &  3.88  \cellcolor{blue!20}  &  &  \\
       $\;\;$ mult-dep-L-     &     &  4.5   \cellcolor{blue!20}  &  &  \\
       $\;\;$ mult-dep-L+     &     &  4.64  \cellcolor{blue!20}  &  &  \\
       $\;\;$ mult-dom-L-     &     &  17.42 \cellcolor{blue!20}  &  &  \\
       $\;\;$ mult-dom-L+     &     &  18.29 \cellcolor{blue!20}  &  &  \\
       $\;\;$ mult-emb-L-     &     &  17.05 \cellcolor{blue!20}  &  &  \\
       $\;\;$ mult-emb-L+     &     &  16.25 \cellcolor{blue!20}  &  &  \\
\hline
\multirow{3}{*}{\textbf{Type II}}   & mean $r$    & n/a & 0.577 \cellcolor{blue!20}  & n/a\\
                                    & mean $\rho$ & n/a & 0.571 \cellcolor{blue!20}  & n/a\\
                                    & $W$         & n/a & 0.598\cellcolor{blue!20}  & n/a \\
%\hline
%       \multirow{1}{*}{\textbf{Type III}}  & -- & -- & \cellcolor{blue!20}-- & -- \\
\hline
       \multirow{1}{*}{\textbf{Type IV}}  & $P$ & n/a & 0.661  & \cellcolor{blue!20} 0.661 \\
\hline
       
    \end{tabular}\caption{Example 3: QRA++ reproducibility assessment for eight comparable experiments ($n=8)$ from REPROLANG 2020: QC = \textbf{\textul{weighted F1 score}}; \citet{vajjala-rama-2018-experiments,bestgen-2020-reproducing,caines-buttery-2020-reprolang,huber-coltekin-2020-reproduction,arhiliuc-etal-2020-language}. %Type~III row marked `--' because there were no such results; 
    Study level and QC level columns same for Type I and IV, because only one QC taken into account. For measures, see Table~\ref{tab:res-types} and Section~\ref{sec:repro-measures}; n/a $=$ measure does not apply at this level.}
    \label{tab:example-3}
\end{table*}

\subsection{Example 2: Quality and socio-political acceptability of news headlines in 2 comparable experiments}

Table~\ref{tab:example-1} reports QRA++ results based on two comparable experiments \cite{gabriel-etal-2022-misinfo, repronlp_submission_34_mahlaza} for two QCs: Overall Quality and Socio-political Acceptability. The experiment properties from Section~\ref{sec:exp-prop} were exactly the same in all three experiments. Overall Quality was assessed on a 5-point rating scale; Socio-political Acceptability as the proportion of items judged acceptable.

Reproducibility results here indicate fundamental differences between the two QCs.
For Socio-political Acceptability (QC 2), degree of reproducibility is good to medium in terms of CV$^*$ (Type~I) at the system and QC level, and also in terms of the correlation coefficients (Type~II) and $P$ (Type~IV) at the QC level.
However, for Overall Quality (QC 1), while CV$^*$ shows medium to poor Type I reproducibility at system and QC level, it is as bad as it can be in terms of QC level correlation coefficients (Type~II) and $P$ measure (Type~IV). 

At the study level, where we do not distinguish between the two QCs, all we can say is that overall reproducibility was medium in terms of mean CV$^*$ and medium to poor in terms of the $P$ measure.

\subsection{Example 3: Weighted F1 of automatic essay scoring in 8 comparable experiments}

As our final example, Table~\ref{tab:example-3} reports a QRA++ assessment for a set of eight comparable experiments conducted by five separate groups of researchers \cite{vajjala-rama-2018-experiments,bestgen-2020-reproducing,caines-buttery-2020-reprolang,huber-coltekin-2020-reproduction,arhiliuc-etal-2020-language}.\footnote{The reproduction studies from 2020 were conducted as part of REPROLANG \cite{branco-etal-2020-shared}.} The evaluation method here is an automatic metric, weighted F1 score (wF1). The eleven systems (column~1) are variants of the same multilingual essay scoring system. Note that the experiment properties from Section~\ref{sec:exp-prop} were \textit{not} the same in all eight individual experiments, a point we return to below.

Because the systems form an (incomplete) ablation set (e.g.\ mult-word-L+ uses word n-grams and a language identity feature in the input), this offers unique insight into differences in degree of reproducibility between system types, moreover on the solid basis of 11 systems and eight comparable experiments (a number that is surely unique in NLP). We can see that in terms of Type I results (CV$^*$), the system variants using POS n-grams have the best reproducibility, closely followed by those using dependency n-grams. Next are the two systems using word n-grams. Finally, the four systems using features specific to the automatic essay scoring domain (e.g.\ document length, errors, etc.), or word and character embeddings, have the poorest wF1 reproducibility by some margin. 

Note that presence or absence of the language identity feature (L+/-) alone makes next to no difference, providing confirmatory evidence that the features have their own, separately quantifiable effect on degree of reproducibility.

In contrast to Examples~1 and~2 above, the eight experiments are not all identical in terms of the experiment properties from Section~\ref{sec:exp-prop}. In particular, (i) the \textbf{\textit{test dataset}} outputs; 
(ii) the \textbf{\textit{metric implementation}}; and (iii) the \textbf{\textit{run-time environment}} in which the code was executed, all differed between different subsets of the eight experiments.

These differences in experiment properties mean that we would expect worse reproducibility than we would see if the properties were all the same. In fact, the more properties that are different the worse reproducibility should tend to get. 

With QRA++, we can examine such effects directly. Table~\ref{tab:same-vs-diff-exp-properties} compares two subsets of the experiments from Table~\ref{tab:example-3} in terms of mean CV$^*$, $W$, and $P$. The first subset is the four experiments that share the same test dataset and random seed, shown in subtable~(a); the second subset have test samples and seeds that differ between them as well as the experiments in the first subset, shown in subtable~(b).

\begin{table*}
\small\centering
\begin{tabular}{cc}
     
    \begin{tabular}{|l|c|c|}
        \hline
        \multicolumn{3}{|c|}{\textit{Same dataset and seed}} \\
        \hline
        \textbf{Type of} & \textbf{Measure} & \textbf{Degree of reproducibility}\\
         \textbf{result} & \textbf{applied}&  ($n=4$), study level \\
        \hline
        \hline
        \textbf{Type I} & mean CV$^*$$\downarrow$ & 8.717\\
        \hline
        \textbf{Type II} & $W$$\uparrow$ & 0.712\\
        \hline
        \textbf{Type IV} & $P$$\uparrow$ & 0.706\\ 
        \hline
        \multicolumn{3}{c}{(a)}
    \end{tabular}
&  

    \begin{tabular}{|l|c|c|}
        \hline
        \multicolumn{3}{|c|}{\textit{Different datasets and seeds}} \\
        \hline
        \textbf{Type of} & \textbf{Measure} & \textbf{Degree of reproducibility}\\
        \textbf{result}  & \textbf{applied}&  ($n=4$), study level \\
        \hline
        \hline
        \textbf{Type I} & mean CV$^*$$\downarrow$ & 14.872\\
        \hline
        \textbf{Type II} & $W$$\uparrow$ & 0.509\\
        \hline
        \textbf{Type IV} & $P$$\uparrow$ & 0.558\\ 
        \hline
        \multicolumn{3}{c}{(b)}
    \end{tabular}

\end{tabular}
\caption{Comparison of subsets of comparable experiments from Table~\ref{tab:example-3} with \textbf{\textul{same dataset and seed}} (a), vs.\ \textbf{\textul{different dataset and seeds}} (b).}
\label{tab:same-vs-diff-exp-properties}
\end{table*}

Comparing (a) and (b), we can see that the reproducibility measures are substantially better for the (more) homogeneous set of experiments in (a), and clearly worse for the heterogeneous set, with just 55.8\% of pairwise ranks the same.

Overall, even for (a), reproducibility is worse than it should be for automatic metrics, and this is connected to the fact that the test set outputs are generated in different run-time environments which, as we saw above, leads to bigger differences for some system types (notably those that use language models) than others.

\section{Discussion}\label{sec:disc}

The examples above each illustrate different aspects of degree-of-reproducibility assessment with QRA++, and the insights it facilitates. Example~1 illustrates that the relative assessment of systems can have (near) perfect reproducibility, even where the reproducibility of the absolute assessment of systems is only medium good. 

Example~2 shows why the QC level is the main level of reproducibility assessment in NLP/ML and why it makes sense to treat evaluations of different QCs as separate experiments even if carried out at the same time: there can be considerable differences in reproducibility depending on quality criterion, or more precisely, depending on the evaluation method that has been implemented in a given experiment for the quality criterion. At the same time, as with Example~1, degree of reproducibility measures for different result types can yield different assessments even for the same quality criterion. 

Finally, Example~3 demonstrates that degree of reproducibility can differ for different types of systems, and that reproducibility assessments grounded in experiment similarity can reveal how different experiment properties can affect reproducibility.

It might be argued that it doesn't matter what the reproducibility of individual experiments is, and that the important thing is what the trends are over a number of multiple, independently conducted experiments. In fields like medicine, systematic reviews play an important role in establishing such trends. However, NLP/ML rarely if ever conducts such reviews for research questions as precise and focused as in medicine. Consider what the equivalent would be of \textit{How does influenza vaccination of healthcare workers relate to patient morbidity and mortality?} Work in NLP/ML isn't aligned, focused or possibly numerous enough to allow this type of systematic review. Moreover, medicine still assesses reproducibility of measures. Another aspect is that NLP/ML is far less standardised in its terminology and definitions, experimental design and reporting than other fields of science, including infrequently reporting work in terms of research questions and hypothesis testing.

\section{Conclusion}\label{sec:concl}

This paper has presented QRA++, a quantitative framework for assessing the {degree} of reproducibility of two or more comparable experiments which produces results that are comparable across different reproducibility assessments, and supports conclusions about aspects of experiments associated with good/bad reproducibility. Via three example applications we demonstrated how QRA++ can be used to obtain a rounded picture of degree of reproducibilty at different levels of granularity, but also to examine the impact of different aspects of an experiment on degree of reproducibility. While we don't have a more reliable way of establishing study-independent system performance than `suck it and see,' it seems sensible to strive to make our experimental methods as reproducible as possible. For this we need to be able  both (i) to assess reproducibility quantitatively and comparably, and (ii) to leverage assessment results for improving the methods, abilities provided by QRA++. 

%\section*{Acknowledgments}

\section*{Ethics Statement}

We did not carry out any new human evaluation experiments for the purposes of the research reported in this paper. Nor did we create new systems or datasets. The ethical risk is therefore minimal.

\bibliography{qra++}

\appendix

\renewcommand{\UrlFont}{\ttfamily\small}
\newcommand{\egcattribute}[1]{\textsc{#1}}
\newcommand{\egcvalue}[1]{\textbf{\textit{#1}}}

\vspace{1cm}

\section{HEDS 3.0 Questions Used as Experiment Properties in Table~1}\label{sec:appendix-exp-prop}

In this appendix, we present the full HEDS 3.0 questions and range of answers serving as the experiment properties for QRA++, as listed in Table~\ref{tab:exp-prop}.
E.g.\ H2.1 in Table~\ref{tab:exp-prop} refers Question 2.1 below (the H standing for HEDS).

\subsection*{\qsecboxqtn{Question 2.1: What type of input do the evaluated system(s) take?}}

\vspace{-.1cm}

\noindent \textit{Notes:} The term `input' here refers to the text, representations and/or data structures that all of the evaluated systems take as input (including prompts).  This question is about input \textit{type}, regardless of number. E.g.\  if the input is a set of documents, you would still select `text: document' below.

\vspace{.1cm}
\noindent \textit{Check-box options (select all that apply)}:
\begin{enumerate}[itemsep=0cm,leftmargin=0.5cm]
	\item[\checkbox] \egcvalue{Raw/structured data}:~~Numerical, symbolic, and other data, possibly structured into trees, graphs, graphical models, etc. E.g.\  the input to Referring Expression Generation (REG), end-to-end text generation, etc.  NB: excludes linguistic representations. % Option Number: 1

	\item[\checkbox] \egcvalue{Deep linguistic representation (DLR)}:~~Any of a variety of deep, underspecified, semantic representations, such as abstract meaning representations (AMRs; \citet{banarescu-etal-2013-abstract}) or discourse representation structures (DRSs; \citet{kamp-reyle-2013}). % Option Number: 2

	\item[\checkbox] \egcvalue{Shallow linguistic representation (SLR)}:~~Any of a variety of shallow, syntactic representations, e.g.\  Universal Dependency (UD) structures; typically the input to surface realisation. % Option Number: 3

	\item[\checkbox] \egcvalue{Text: subsentential unit of text}:~~Unit(s) of text shorter than a sentence, e.g.\  Referring Expressions (REs), verb phrase, text fragment of any length; includes titles/headlines. % Option Number: 4

	\item[\checkbox] \egcvalue{Text: sentence}:~~Single sentence(s). % Option Number: 5

	\item[\checkbox] \egcvalue{Text: multiple sentences}:~~Sequence(s) of multiple sentences, without any document structure. % Option Number: 6

	\item[\checkbox] \egcvalue{Text: document}:~~Text(s) with document structure, such as a title, paragraph breaks or sections, e.g.\  a set of news reports for summarisation. % Option Number: 7

	\item[\checkbox] \egcvalue{Text: dialogue}:~~Dialogue(s) of any length, excluding a single turn which would come under one of the other text types. % Option Number: 8

	\item[\checkbox] \egcvalue{Text: other (please describe)}:~~Input is text but doesn't match any of the above text categories. % Option Number: 9

	\item[\checkbox] \egcvalue{Speech}:~~Recording(s) of speech. % Option Number: 10

	\item[\checkbox] \egcvalue{Visual}:~~Image(s) or video(s). % Option Number: 11

	\item[\checkbox] \egcvalue{Multi-modal}:~~Select this option if input is\textit{always} a combination of multiple modalities. Also select other options in this list to different elements of the multi-modal input. % Option Number: 12

	\item[\checkbox] \egcvalue{Control feature}:~~Feature(s) or parameter(s) specifically present to control a property of the output text, e.g.\  positive stance, formality, author style. % Option Number: 13

	\item[\checkbox] \egcvalue{No input (please explain)}:~~If there are no system inputs, select this option and explain why. % Option Number: 14

	\item[\checkbox] \egcvalue{Other (please describe)}:~~If input is none of the above, select this option and describe it. % Option Number: 15

\end{enumerate}

\subsection*{\qsecboxqtn{Question 2.2: What type of output do the evaluated system(s) generate?}}

\noindent \textit{Notes:} The term `output' here refers to the text, representations and/or data structures that all of the evaluated systems produce as output. This question is about output type, regardless of number. E.g.\  if the output is a set of documents, you would still select `text: document' below.

\vspace{.1cm}
\noindent \textit{Check-box options (select all that apply)}:
\begin{enumerate}[itemsep=0.1cm,leftmargin=0.5cm]
	\item[\checkbox] \egcvalue{Raw/structured data}:~~Numerical, symbolic, and other data, possibly structured into trees, graphs, graphical models, etc. E.g.\  the input to Referring Expression Generation (REG), end-to-end text generation, etc.  NB: excludes linguistic representations. % Option Number: 1

	\item[\checkbox] \egcvalue{Deep linguistic representation (DLR)}:~~Any of a variety of deep, underspecified, semantic representations, such as abstract meaning representations (AMRs; \citet{banarescu-etal-2013-abstract}) or discourse representation structures (DRSs; \citet{kamp-reyle-2013}). % Option Number: 2

	\item[\checkbox] \egcvalue{Shallow linguistic representation (SLR)}:~~Any of a variety of shallow, syntactic representations, e.g.\  Universal Dependency (UD) structures; typically the input to surface realisation. % Option Number: 3

	\item[\checkbox] \egcvalue{Text: subsentential unit of text}:~~Unit(s) of text shorter than a sentence, e.g.\  Referring Expressions (REs), verb phrase, text fragment of any length; includes titles/headlines. % Option Number: 4

	\item[\checkbox] \egcvalue{Text: sentence}:~~Single sentence(s). % Option Number: 5

	\item[\checkbox] \egcvalue{Text: multiple sentences}:~~Sequence(s) of multiple sentences, without any document structure. % Option Number: 6

	\item[\checkbox] \egcvalue{Text: document}:~~Text(s) with document structure, such as a title, paragraph breaks or sections, e.g.\  a set of news reports for summarisation. % Option Number: 7

	\item[\checkbox] \egcvalue{Text: dialogue}:~~Dialogue(s) of any length, excluding a single turn which would come under one of the other text types. % Option Number: 8

	\item[\checkbox] \egcvalue{Text: other (please describe)}:~~Input is text but doesn't match any of the above text categories. % Option Number: 9

	\item[\checkbox] \egcvalue{Speech}:~~Recording(s) of speech. % Option Number: 10

	\item[\checkbox] \egcvalue{Visual}:~~Image(s) or video(s). % Option Number: 11

	\item[\checkbox] \egcvalue{Multi-modal}:~~Select this option if input is\textit{always} a combination of multiple modalities. Also select other options in this list to different elements of the multi-modal input. % Option Number: 12

	\item[\checkbox] \egcvalue{No input (please explain)}:~~If there are no system inputs, select this option and explain why. % Option Number: 14

	\item[\checkbox] \egcvalue{Other (please describe)}:~~If input is none of the above, select this option and describe it. % Option Number: 15

\end{enumerate}

\subsection*{\qsecboxqtn{Question 2.3: What is the task that the evaluated system(s) perform in mapping the inputs in Question~2.1 to the outputs in Question~2.2?}}

\noindent \textit{Notes:} This question is about the task(s) performed by the system(s) being evaluated.  This is independent of the application domain (financial reporting, weather forecasting, etc.), or the specific method (rule-based, neural, etc.) implemented in the system. We indicate mutual constraints between inputs, outputs and task for some of the options below.

\vspace{.1cm}
\noindent \textit{Check-box options (select all that apply)}:
\begin{enumerate}[itemsep=0cm,leftmargin=0.5cm]
	\item[\checkbox] \egcvalue{Content selection/determination}:~~Selecting the specific content that will be expressed in the generated text from a representation of possible content. This could be attribute selection for REG (without the surface realisation step). Note that the output here is not text. % Option Number: 1

	\item[\checkbox] \egcvalue{Content ordering/structuring}:~~Assigning an order and/or structure to content to be included in generated text. Note that the output here is not text. % Option Number: 2

	\item[\checkbox] \egcvalue{Aggregation}:~~Converting inputs (typically \textit{deep linguistic representations} or \textit{shallow linguistic representations}) in some way in order to reduce redundancy (e.g.\  representations for `they like swimming', `they like running' → representation for `they like swimming and running'). % Option Number: 3

	\item[\checkbox] \egcvalue{Referring expression generation}:~~Generating \textit{text} to refer to a given referent, typically represented in the input as a set of attributes or a linguistic representation. % Option Number: 4

	\item[\checkbox] \egcvalue{Lexicalisation}:~~Associating (parts of) an input representation with specific lexical items to be used in their realisation. % Option Number: 5

	\item[\checkbox] \egcvalue{Deep generation}:~~One-step text generation from \textit{raw/structured data} or \textit{deep linguistic representations}. One-step means that no intermediate representations are passed from one independently run module to another. % Option Number: 6

	\item[\checkbox] \egcvalue{Surface realisation (SLR to text)}:~~One-step text generation from \textit{shallow linguistic representations}. One-step means that no intermediate representations are passed from one independently run module to another. % Option Number: 7

	\item[\checkbox] \egcvalue{Feature-controlled text generation}:~~Generation of text that varies along specific dimensions where the variation is controlled via \textit{control features} specified as part of the input. Input is a non-textual representation (for feature-controlled text-to-text generation select the matching text-to-text task). % Option Number: 8

	\item[\checkbox] \egcvalue{Data-to-text generation}:~~Generation from \textit{raw/structured data} which may or may not include some amount of content selection as part of the generation process. Output is likely to be \textit{text:} or \textit{multi-modal}. % Option Number: 9

	\item[\checkbox] \egcvalue{Dialogue turn generation}:~~Generating a dialogue turn (can be a greeting or closing) from a representation of dialogue state and/or last turn(s), etc. % Option Number: 10

	\item[\checkbox] \egcvalue{Question generation}:~~Generation of questions from given input text and/or knowledge base such that the question can be answered from the input. % Option Number: 11

	\item[\checkbox] \egcvalue{Question answering}:~~Input is a question plus optionally a set of reference texts and/or knowledge base, and the output is the answer to the question. % Option Number: 12

	\item[\checkbox] \egcvalue{Paraphrasing/lossless simplification}:~~Text-to-text generation where the aim is to preserve the meaning of the input while changing its wording. This can include the aim of changing the text on a given dimension, e.g.\  making it simpler, changing its stance or sentiment, etc., which may be controllable via input features. Note that this task type includes meaning-preserving text simplification (non-meaning preserving simplification comes under \textit{compression/lossy simplification} below). % Option Number: 13

	\item[\checkbox] \egcvalue{Compression/lossy simplification}:~~Text-to-text generation that has the aim to generate a shorter, or shorter and simpler, version of the input text. This will normally affect meaning to some extent, but as a side effect, rather than the primary aim, as is the case in \textit{summarisation}. % Option Number: 14

	\item[\checkbox] \egcvalue{Machine translation}:~~Translating text in a source language to text in a target language while maximally preserving the meaning. % Option Number: 15

	\item[\checkbox] \egcvalue{Summarisation (text-to-text)}:~~Output is an extractive or abstractive summary of the important/relevant/salient content of the input document(s). % Option Number: 16

	\item[\checkbox] \egcvalue{End-to-end text generation}:~~Use this option if the system task corresponds to more than one of tasks above, but the system doesn't implement them as separate tasks. % Option Number: 17

	\item[\checkbox] \egcvalue{Image/video description}:~~Input includes \textit{visual}, and the output describes it in some way. % Option Number: 18

	\item[\checkbox] \egcvalue{Post-editing/correction}:~~The system edits and/or corrects the input text (can itself be the textual output from another system) to yield an improved version of the text. % Option Number: 19

	\item[\checkbox] \egcvalue{Other (please describe)}:~~If task is none of the above, Select this option and describe it. % Option Number: 20

\end{enumerate}

\subsection*{\qsecboxqtn{Question 3.1.1: How many system outputs (or other evaluation items) are evaluated per system?}}
\vspace{-.1cm}

\noindent \textit{What to enter in the text box}:~~The number of system outputs (or other evaluation items) that are evaluated per system by at least one evaluator in the experiment.  For most experiments this should be a single integer.  If the number of outputs varies please explain how and why.

\subsection*{\qsecboxqtn{Question 4.2.1: Does an individual assessment involve an objective or a subjective judgment?}}

\vspace{-.1cm}
\noindent \textit{Multiple-choice options (select one)}:
\vspace{-.1cm}
\begin{enumerate}[itemsep=0cm,leftmargin=0.5cm]
	\item[\radiobutton] \egcvalue{Objective}:~~Select this option if the evaluation uses objective assessment, e.g.\  any automatically counted or otherwise quantified measurements such as mouse-clicks, occurrences in text, etc. Repeated assessments of the same output with an objective-mode evaluation method should yield the same score/result. % Option Number: 1

	\item[\radiobutton] \egcvalue{Subjective}:~~Select this option in all other cases. Subjective assessments involve ratings, opinions and preferences by evaluators. Some criteria lend themselves more readily to subjective assessments, e.g.\  Friendliness of a conversational agent, but an objective measure e.g.\  based on lexical markers is also conceivable. % Option Number: 2

\end{enumerate}

\subsection*{\qsecboxqtn{Question 4.2.2: Are outputs assessed in absolute or relative terms?}}

\vspace{.1cm}
\noindent \textit{Multiple-choice options (select one)}:
\begin{enumerate}[itemsep=0cm,leftmargin=0.5cm]
	\item[\radiobutton] \egcvalue{Absolute}:~~Select this option if evaluators are shown outputs from a single system during each individual assessment. % Option Number: 1

	\item[\radiobutton] \egcvalue{Relative}:~~Select this option if evaluators are shown outputs from multiple systems at the same time during assessments, typically ranking or preference-judging them. % Option Number: 2

\end{enumerate}

\subsection*{\qsecboxqtn{Question 4.2.3: Is the evaluation intrinsic or extrinsic?}}

\vspace{.1cm}
\noindent \textit{Multiple-choice options (select one)}:
\begin{enumerate}[itemsep=0cm,leftmargin=0.5cm]
	\item[\radiobutton] \egcvalue{Intrinsic}:~~Select this option if quality of outputs is assessed \textit{without} considering their effect on something external to the system such as the performance of an embedding system or of a user at a task. % Option Number: 1

	\item[\radiobutton] \egcvalue{Extrinsic}:~~Select this option if quality of outputs \textit{is} assessed in terms of their effect on something external to the system such as the performance of an embedding system or of a user at a task. % Option Number: 2

\end{enumerate}

\subsection*{\qsecboxqtn{Question 3.2.1: How many evaluators are there in this experiment?}}
\vspace{-.1cm}

\noindent \textit{What to enter in the text box}:~~A single integer representing the total number of evaluators whose assessments contribute to results in the experiment. Don't count evaluators who performed some evaluations but who were subsequently excluded.
\vspace{.1cm}

\subsection*{\qsecboxqtn{Question 3.2.2.1: Are the evaluators in this experiment domain experts?}}

\vspace{-.1cm}
\noindent \textit{Multiple-choice options (select one)}:
\begin{enumerate}[itemsep=0cm,leftmargin=0.5cm]
	\item[\radiobutton] \egcvalue{Yes}:~~Participants are considered domain experts, e.g.\  meteorologists evaluating a weather forecast generator, or nurses evaluating an ICU report generator. % Option Number: 1

	\item[\radiobutton] \egcvalue{No}:~~Participants are not domain experts. % Option Number: 2

	\item[\radiobutton] \egcvalue{N/A (please explain).} % Option Number: 3

\end{enumerate}

\subsection*{\qsecboxqtn{Question 3.2.2.4: Were any of the researchers running the experiment among the participants?}}

\vspace{-.15cm}
\noindent \textit{Multiple-choice options (select one)}:
\begin{enumerate}[itemsep=-0.1cm,leftmargin=0.5cm]
	\item[\radiobutton] \egcvalue{Yes}:~~Evaluators include one or more of the researchers running the experiment. % Option Number: 1

	\item[\radiobutton] \egcvalue{No}:~~Evaluators do not include any of the researchers running the experiment. % Option Number: 2

	\item[\radiobutton] \egcvalue{N/A (please explain).} % Option Number: 3

\end{enumerate}

\subsection*{\qsecboxqtn{Question 3.2.4: What training and/or practice are evaluators given before starting on the evaluation itself?}}

\vspace{-.15cm}
\noindent \textit{What to enter in the text box}:~~Describe any training evaluators were given to prepare them for the evaluation task, including any practice evaluations they did. This includes introductory explanations, e.g.\  on the start page of an online evaluation tool.

\subsection*{\qsecboxqtn{Question 3.2.5: What other characteristics do the evaluators have?}}

\vspace{-.15cm}
\noindent \textit{What to enter in the text box}:~~Use this space to list any characteristics not covered in previous questions that the evaluators are known to have, e.g.\  because of information collected during the evaluation. This might include geographic location, educational level, or demographic information such as gender, age, etc. Where characteristics differ among evaluators (e.g.\  gender, age, location etc.), also give numbers for each subgroup.

\subsection*{\qsecboxqtn{Question 3.3.2: By what medium are responses collected?}}

\vspace{-.1cm}
\noindent \textit{What to enter in the text box}:~~Describe the platform or other medium used to collect responses, e.g.\  paper forms, Google forms, SurveyMonkey, Mechanical Turk, CrowdFlower, audio/video recording, etc.

\subsection*{\qsecboxqtn{Question 3.3.3.1: What types of quality assurance methods are used to ensure that evaluators are sufficently qualified and/or their responses are of sufficient quality?}}

\noindent If any quality assurance methods other than those listed were used, select `other', and describe why below.  If no methods were used, select \textit{none of the above}.

\vspace{.1cm}
\noindent \textit{Check-box options (select all that apply)}:
\begin{enumerate}[itemsep=0cm,leftmargin=0.5cm]
	\item[\checkbox] \egcvalue{Evaluators are required to be native speakers of the language they evaluate}:~~Mechanisms are in place to ensure all participants are native speakers of the language they evaluate. % Option Number: 1

	\item[\checkbox] \egcvalue{Automatic quality checking methods are used during and/or after evaluation}:~~Evaluations are checked for quality by automatic scripts during or after evaluations, e.g.\  evaluators are given known bad/good outputs to check that scores are appropriate. % Option Number: 2

	\item[\checkbox] \egcvalue{Manual quality checking methods are used during/post evaluation}:~~Evaluations are checked for quality by a manual process during or after evaluations, e.g.\  scores assigned by evaluators are monitored by researchers conducting the experiment. % Option Number: 3

	\item[\checkbox] \egcvalue{Evaluators are excluded if they fail quality checks (often or badly enough)}:~~There are conditions under which evaluations produced by participants are not included in the final results due to quality issues. % Option Number: 4

	\item[\checkbox] \egcvalue{Some evaluations are excluded because of failed quality checks}:~~There are conditions under which some (but not all) of the evaluations produced by some participants are not included in the final results due to quality issues. % Option Number: 5

	\item[\checkbox] \egcvalue{Other (please describe)}:~~Briefly mention any other quality-assurance methods that were used. Details of the method should be entered under 3.3.3.2. % Option Number: 6

	\item[\checkbox] \egcvalue{None of the above (no quality assurance methods used).} % Option Number: 7

\end{enumerate}

\subsection*{\qsecboxqtn{Question 4.3.1.2: What standardised quality criterion name does the name entered for 4.3.1.1 correspond to?}}

\noindent \textit{What to enter in the text box}:~~Map the quality criterion name used in the evaluation experiment to its equivalent in a standardised set of quality criterion names and definitions such as QCET \citep{belz-etal-2024-qcet-interactive, belz-etal-2025-qcet}, and enter the standardised name and reference to the paper here. In performing this mapping, the information given in Questions 4.3.7 (question/prompt), 3.3.4.1--3.3.4.2 (interface/information shown to evaluators), 4.3.2 (QC definition), 3.2.4 (training/practice), and 4.3.1.1 (verbatim QC name) should be taken into account, in this order of precedence.

\subsection*{\qsecboxqtn{Question 4.3.5: How is the scale or other rating instrument presented to evaluators?}}

%\vspace{.1cm}
\noindent \textit{Multiple-choice options (select one)}:
\begin{enumerate}[itemsep=0cm,leftmargin=0.5cm]
	\item[\radiobutton] \egcvalue{Multiple-choice options}:~~Select this option if evaluators select exactly one of multiple options. % Option Number: 1

	\item[\radiobutton] \egcvalue{Check-boxes}:~~Select this option if evaluators select any number of options from multiple given options. % Option Number: 2

	\item[\radiobutton] \egcvalue{Slider}:~~Select this option if evaluators move a pointer on a slider scale to the position corresponding to their assessment. % Option Number: 3

	\item[\radiobutton] \egcvalue{N/A (there is no rating instrument)}:~~Select this option if there is no rating instrument. % Option Number: 4

	\item[\radiobutton] \egcvalue{Other (please describe)}:~~Select this option if there is a rating instrument, but none of the above adequately describe the way you present it to evaluators. Use the text box to describe the rating instrument and link to a screenshot. % Option Number: 5

\end{enumerate}

\subsection*{\qsecboxqtn{Question 4.3.7: What is the verbatim question, prompt or instruction given to evaluators (visible to them during each individual assessment)?}}

\noindent \textit{What to enter in the text box}:~~Copy and paste the verbatim text that evaluators see during each assessment, that is intended to convey the evaluation task to them. E.g.\  \textit{Which of these texts do you prefer?} Or \textit{Make any corrections to this text that you think are necessary in order to improve it to the point where you would be happy to provide it to a client.}

\subsection*{\qsecboxqtn{Question 4.3.8: What form of response elicitation is used in collecting assessments from evaluators?}}

\noindent The terms and explanations in this section have been adapted from \citet{howcroft-etal-2020-twenty}.

\vspace{.2cm}
\noindent \textit{Multiple-choice options (select one)}:
\begin{enumerate}[itemsep=0cm,leftmargin=0.5cm]
	\item[\radiobutton] \egcvalue{(Dis)agreement with quality statement}:~~Participants indicate the degree to which they agree with a given quality statement on a rating instrument. The rating instrument is labelled with degrees of agreement and can additionally have numerical labels. E.g.\ \textit{This text is fluent: 1=strongly disagree\ldots5=strongly agree}. % Option Number: 1

	\item[\radiobutton] \egcvalue{Direct quality estimation}:~~Participants indicate level of quality on a rating instrument, which typically (but not always) mentions the quality criterion explicitly. E.g.\  \textit{How fluent is this text? 1=not at all fluent\ldots5=very fluent}. % Option Number: 2

	\item[\radiobutton] \egcvalue{Relative quality estimation (including ranking)}:~~Participants evaluate two or more items in terms of which is better. E.g.\  \textit{Rank these texts in terms of Fluency}: \textit{Which of these texts is more fluent?} \textit{Which of these items do you prefer?} % Option Number: 3

	\item[\radiobutton] \egcvalue{Counting occurrences in text}:~~Evaluators are asked to count how many times some type of phenomenon occurs, e.g.\  the number of facts contained in the output that are inconsistent with the input. % Option Number: 4

	\item[\radiobutton] \egcvalue{Qualitative feedback (e.g.\  via comments entered in a text box)}:~~Typically, these are responses to open-ended questions in a survey or interview. % Option Number: 5

	\item[\radiobutton] \egcvalue{Evaluation through post-editing/ annotation}:~~Select this option if the evaluators' task consists of editing, or inserting annotations in, text. E.g.\  evaluators may perform error correction and edits are then automatically measured to yield a numerical score. % Option Number: 6

	\item[\radiobutton] \egcvalue{Output classification or labelling}:~~Select this option if evaluators assign outputs to categories. E.g.\  \textit{What is the overall sentiment of this piece of text? — Positive/neutral/negative.} % Option Number: 7

	\item[\radiobutton] \egcvalue{User-text interaction measurements}:~~Select this option if participants in the evaluation experiment interact with a text in some way, and measurements are taken of their interaction. E.g.\  reading speed, eye movement tracking, comprehension questions, etc. Excludes situations where participants are given a task to solve and their performance is measured which comes under the next option. % Option Number: 8

	\item[\radiobutton] \egcvalue{Task performance measurements}:~~Select this option if participants in the evaluation experiment are given a task to perform, and measurements are taken of their performance at the task. E.g.\  task is finding information, and task performance measurement is task completion speed and success rate. % Option Number: 9

	\item[\radiobutton] \egcvalue{User-system interaction measurements}:~~Select this option if participants in the evaluation experiment interact with a system in some way, while measurements are taken of their interaction. E.g.\  duration of interaction, hyperlinks followed, number of likes, or completed sales. % Option Number: 10

	\item[\radiobutton] \egcvalue{Other (please describe)}:~~Use the text box to describe the form of response elicitation used in assessing the quality criterion if it doesn't fall in any of the above categories. % Option Number: 11

\end{enumerate}

\end{document}